%% file: main.tex
\documentclass{article}
\usepackage{spconf,amsmath,graphicx}
\usepackage{amssymb}
\usepackage{booktabs}
\usepackage{multirow}
\usepackage{xcolor}
\usepackage{xspace}
\usepackage{changes}

\newcommand{\mname}{LoCoMT\xspace}

\title{Multimodal Transformer with a Low-Computational-Cost Guarantee}
\name{Sungjin Park \qquad Edward Choi}
\address{KAIST}

\begin{document}
\ninept
\maketitle
\begin{abstract}
\vspace{0.1mm}
\input{contents/00Abstract}
\end{abstract}


\section{Introduction}
\label{sec:intro}
\input{contents/01Introduction}

\section{Methods}
\label{sec:methods}
\input{contents/02Methods}

\section{Experiments}
\label{sec:exp}
\input{contents/03Experiments}

\section{Conclusion}
\label{sec:conclusion}
\input{contents/04Conclusion}

\section{Acknowledgement}
\label{sec:acknowl}
This work was supported by the Institute of Information \& Communications Technology Planning \& Evaluation (IITP) grant (No.2019-0-00075, No.2022-0-00984), and National Research Foundation of Korea (NRF) grant (NRF-2020H1D3A2A03100945), funded by the Korea government (MSIT).

\bibliographystyle{IEEEbib}
\bibliography{refs}

\end{document}

%% file: contents/00Abstract.tex
Transformer-based models have significantly improved performance across a range of multimodal understanding tasks, such as visual question answering~\cite{vqa, vqa2} and action recognition~\cite{kinetics, youtube8M, epic}. 
However, multimodal Transformers significantly suffer from a quadratic complexity of the multi-head attention with the input sequence length, especially as the number of modalities increases.
To address this, we introduce Low-Cost Multimodal Transformer (\mname), a novel multimodal attention mechanism that aims to reduce computational cost during training and inference with minimal performance loss.
Specifically, by assigning different multimodal attention patterns to each attention head, \mname can flexibly control multimodal signals and theoretically ensures a reduced computational cost compared to existing multimodal Transformer variants.
Experimental results on two multimodal datasets, namely Audioset and MedVidCL demonstrate that \mname not only reduces GFLOPs but also matches or even outperforms established models.

%% file: contents/01Introduction.tex
Transformer~\cite{transformer} has become a de facto backbone for multimodal learning, thanks to its flexibility~\cite{perceiver, universalTransformer} and superior performance~\cite{beit3, pali, ofa}. 
One significant limitation, however, is the computational cost of multi-head attention, which scales quadratically with the input sequence length. 
Previous studies have developed efficient Transformers to reduce the computational complexity of the attention mechanism.
Several approaches compute attention scores between a subset of query-key pairs, resulting in a sparse attention matrix. 
Specifically, they used fixed patterns~\cite{blockwise, sparsetransformer, longformer} or learnable patterns~\cite{reformer, routingformer} to sparsify the attention matrix. 
Another emerging type of efficient transformer is to improve efficiency by leveraging the approximation of the attention matrix~\cite{linformer, performer}.
However, when these efficient methods are applied to multimodal Transformers, new challenges arise. 

Multimodal Transformers typically encode multimodal inputs by employing different combinations of attention mechanisms (Fig.~\ref{fig:pattern}), such as early-fusion (\textit{i.e.}, multimodal attention from the first layer to the last), mid-fusion (\textit{i.e.}, self-attention until the middle layer, then multimodal attention or cross-attention to the last), and late-fusion (\textit{i.e.}, self-attention until the last layer, then concatenation of each modality-specific representation vectors).
This diversity in attention types makes addressing their computational complexity a challenge. 
Moreover, the interplay and relevance between different modalities, and their collective contribution to the final prediction, are not always straightforward, especially when compared to unimodal inputs.
The recently proposed Multimodal Bottleneck Transformer (MBT)~\cite{mbt} aims to reduce computational cost by introducing bottleneck tokens into the input sequence. 
In particular, MBT reduces the effective input length of the attention mechanism by allowing interaction between modalities only through the few bottleneck tokens. 
Yet, as we will discuss in Sec.~\ref{sec:methods}, MBT still requires more computations than existing late-fusion models~\cite{multisensory2018}.

\input{figures/pattern}

In this work, we introduce Low-Cost Multimodal Transformer (\mname), a novel attention mechanism for efficient multimodal fusion in Transformer.
Each attention head in \mname computes an attention matrix based on one of the predefined patterns by limiting the set of keys to which queries can refer.
Specifically, we consider two patterns commonly employed in multimodal Transformers: self-attention and cross-attention (Fig.~\ref{fig:pattern}). 
While self-attention extracts the contextualized representation of inputs belonging to the same modality, cross-attention captures the cross-modal interaction between two different modalities.
Through theoretical analysis, we established that \mname reduces the attention cost in proportion to the square of the difference between the sequence lengths of the modalities.
Consequently, \mname not only processes inputs more efficiently than conventional multimodal Transformers, but it also becomes more efficient for multimodal datasets where sequence lengths vary greatly between modalities, such as in video classification.

We evaluated \mname on two datasets: Audioset, an audio-video classification dataset~\cite{audioset}, and MedVidCL, an audio-video-text classification dataset~\cite{medvidcl}. As per standard practice~\cite{mbt, medvidcl, ama}, the number of video tokens in both datasets is larger than that of audio and text tokens.
Our empirical results show that \mname indeed reduces computational cost, while achieving comparable or superior performance compared to existing multimodal Transformers in both datasets.
We further investigated the trade-off between performance and efficiency with various configurations of \mname.
As shown in Fig.~\ref{fig:ablations}, \mname can save GFLOPs by nearly half while still maintaining performance on par with the most expensive baseline.
Additionally, we explored the impact of assigning different attention patterns across layers.
We found that self-attention in low-level layers helps increase performance, while in high-level layers, even a random attention pattern works well.

%% file: figures/pattern.tex
\begin{figure}[t]
    \centering
    \includegraphics[width=1.0\columnwidth]{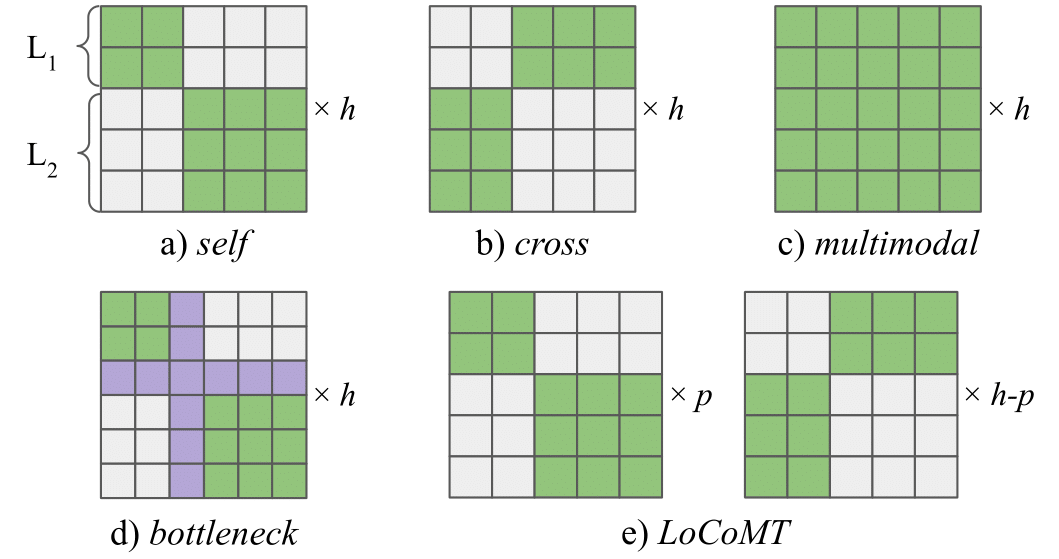}
    \caption{
        Attention map of a)-d) common multimodal Transformers e) \mname when the input consists of two modalities with length $\mathrm{L}_1$ and $\mathrm{L}_2$ and the number of attention head is $h$. We indicate masked tokens as gray, active tokens as green, and bottleneck tokens as purple. We denote the number of attention view $P_i$ assigned to the attention head by $p_i$.
    }
    \label{fig:pattern}
\end{figure}

%% file: contents/02Methods.tex
\input{figures/eco}
We begin with a summary of the standard multi-head attention~\cite{transformer}.
Afterwards, we compare the computational cost of the common attention patterns used in multimodal Transformers. 
Finally, we introduce Low-Cost Multimodal Transformer (\mname), a novel attention mechanism that can further reduce the cost compared to existing attention patterns.

\subsection{Preliminary: Multi-head Attention}
Multi-head attention (MHA)~\cite{transformer} linearly projects queries, keys, and values $n_h$ times (\textit{i.e.,} number of attention heads) with learned linear projections and applies the scaled dot product to the projected features in parallel.
The scaled dot product receives queries and keys of dimension $d_k$ and values of dimension $d_v$ as input and then calculates the dot product of the query with all keys.
In practice, we pack all queries, keys, and values into the matrix $Q \in \mathbb{R}^{L_q \times d_k}$, $K \in \mathbb{R}^{L_{kv} \times d_k}$, and $V \in \mathbb{R}^{L_{kv} \times d_v}$ and apply multi-head attention.

The computational cost $C$ of the scaled-dot product is
\begin{equation}
    C = L_qL_{kv}d_k,
\end{equation}
where $L_q$ is the sequence length of queries, $L_{kv}$ is the sequence length of keys and values. This becomes the most expensive operation in the Transformer when $L_q, L_{kv} \gg d_k$.

\subsection{Common Attention Patterns in Multimodal Transformer}
We analyze the computational cost of four attention patterns (Fig.~\ref{fig:pattern} a)-d) deployed in the multimodal Transformer, specifically the cost of query-key matrix multiplication in a single layer.
For simplicity in cost computation, we assume two input modalities with lengths of $L_1$ and $L_2$, respectively. We also assume that the hidden dimensions for the query, key, and value are all $d$.
Cross-modal encoders in previous studies~\cite{lxmert, vilbert} consecutively applied self-attention and cross-attention to input.
Specifically, self-attention first updates the unimodal representation, and then cross-attention merges information from other modalities. 
The associated computational costs $C_{self}$ and $C_{cross}$ are, respectively,
\begin{equation}
C_{self} = \left(L_1^2+L_2^2\right)d \qquad C_{cross} = 2L_1L_2d.
\end{equation}
The multimodal attention mechanism~\cite{vilt}, another common attention pattern in the multimodal Transformer, allows each token to refer to any other tokens in the sequence. 
However, its computational cost, $C_{multi}$,
\begin{equation}
C_{multi} = \left(L_1+L_2\right)^2d
\end{equation}
is larger than both self and cross attention.

The bottleneck attention was proposed in Multimodal Bottleneck Transformer (MBT)~\cite{mbt} to reduce the cost of multimodal attention.
The core component of the bottleneck attention is a small set of bottleneck tokens, where multimodal interaction occurs only through this.
During the forward process, MBT creates copies of bottleneck tokens that are contextualized separately with different modalities.
Then MBT takes the mean of copies to get the updated representation of bottleneck tokens.
The computational cost $C_{bottle}$ of bottleneck attention is
\begin{equation}
C_{bottle} = \left\{\left(L_1+B\right)^2+\left(L_2+B\right)^2\right\}d
\end{equation}
where $B$ is the number of bottleneck tokens.
In summary, the computational costs follow the relationship:
\begin{equation}
C_{cross} \leq C_{self} < C_{bottle} < C_{multi}
\end{equation}
when $B \ll L_1, L_2$.
The use of bottleneck tokens in bottleneck attention reduces its cost compared to multimodal attention, but it still incurs a higher cost than self-attention.

\subsection{Low-Cost Multimodal Transformer}
We introduce a new attention mechanism, namely Low-Cost Multimodal Transformer (\mname), that enables multimodal fusion at a lower computational cost than self-attention.
By limiting the input modalities to which each attention head can refer, \mname can efficiently compute the attention matrix.
The constraint that we apply to each attention head is denoted as the attention view, which is chosen from the predefined set $P=\{P_1, ..., P_{{}_m C_2+1}\}$, where $m$ is the number of input modalities.
Specifically, $P_1$ represents self-attention, while others, $P_i, i \in [2, {}_m C_2+1]$, facilitate cross-attention between pairs of $m$ input modalities.
For example, with image, audio and text as input, $P_1$ is self-attention, and $P_2$, $P_3$, and $P_4$ are cross-attention between image-text, image-audio, and audio-text, respectively.
We assign a view $P_i$ to each attention head during initialization, which remains fixed for training and inference.
Note that we have the freedom to choose which views are prominently assigned to attention heads and apply these allocations variably across layers, as depicted in Fig.~\ref{fig:eco}.
For example, $P_2$ can be assigned to more heads to emphasize the interaction between the first and second modalities in Fig.~\ref{fig:pattern}.
Owing to this design, \mname enables flexible control of multimodal signals and we investigate its effectiveness in Sec.~\ref{sec:analysis}.
Formally, we compute the output $\mathbf{o}$ of \mname as follows:
\begin{equation}
    \label{eq:locomt}
    \begin{split}
        & \mathbf{o}=\mathrm{FFN}(\mathbf{x}) \quad \mathbf{x} = [\mathbf{h_1};...;\mathbf{h_{n_h}}]W_O \\
        & \mathbf{h_i} = \Bigl(a\bigl(Q^{(i)}_j,K^{(i)},V^{(i)}, P^{(i)}_j\bigr)\Bigr)_{j \in \{1,...,\sum_{k}L_k\}} \\
        & a\bigl(Q^{(i)}_j,K^{(i)},V^{(i)},P^{(i)}_j\bigr) = \sigma\left(\frac{Q^{(i)}_j{K^{(i)}_{*}}^T}{\sqrt{d_k}}\right)V^{(i)}_{*} \\
        & K^{(i)}_{*} = \left(K^{(i)}_l\right)_{l \in P^{(i)}_j} \;\; V^{(i)}_{*} = \left(V^{(i)}_l\right)_{l \in P^{(i)}_j}
    \end{split}
\end{equation}
where $W_O$ is the linear projection for the output, $P^{(i)} \in P$ is the attention view assigned to the head $i$, and $P_j^{(i)}$ consists of indices of the keys and values to which the $j$-th query vector attends.

This design choice effectively reduces the cost $C_\mathrm{\mname}$ of \mname to:
\begin{equation}
C_{\mathrm{\mname}} = \frac{p_0}{n_h}\sum_{i=1}^{m}L_i^2+2\sum_{i=2}^{m}\sum_{j=1}^{i-1}\frac{p_{ij}}{n_h}L_iL_j 
\end{equation}
where $p_0+\sum_{i=2}^{m}\sum_{j=1}^{i-1}p_{ij}=n_h$.
The difference between the cost of self-attention and \mname $C_{self}-C_\mathrm{\mname}$ satisfies the inequality:
\begin{equation*}
    \resizebox{1.0\columnwidth}{!}{$
        C_{self}-C_{\mathrm{\mname}}=\sum_{i=2}^{m}\sum_{j=1}^{i-1}\frac{p_{ij}}{n_h}\left\{\sum_{k=1, k \neq i,j}^{m}L_k^2+(L_i-L_j)^2\right\} \geq 0,
    $}
\end{equation*}
leading to the subsequent inequality:
\begin{equation}
    \begin{split}
        C_\mathrm{\mname} \leq C_{self} < C_{bottle} < C_{multi}
    \end{split}
\end{equation}
Since the equality holds if and only if $m=2$ and $L_1=L_2$, \mname achieves the lower computational cost compared to self-attention, while also providing adaptive control of multimodal fusion.

%% file: figures/eco.tex
\begin{figure}[t]
    \centering
    \includegraphics[width=0.98\columnwidth]{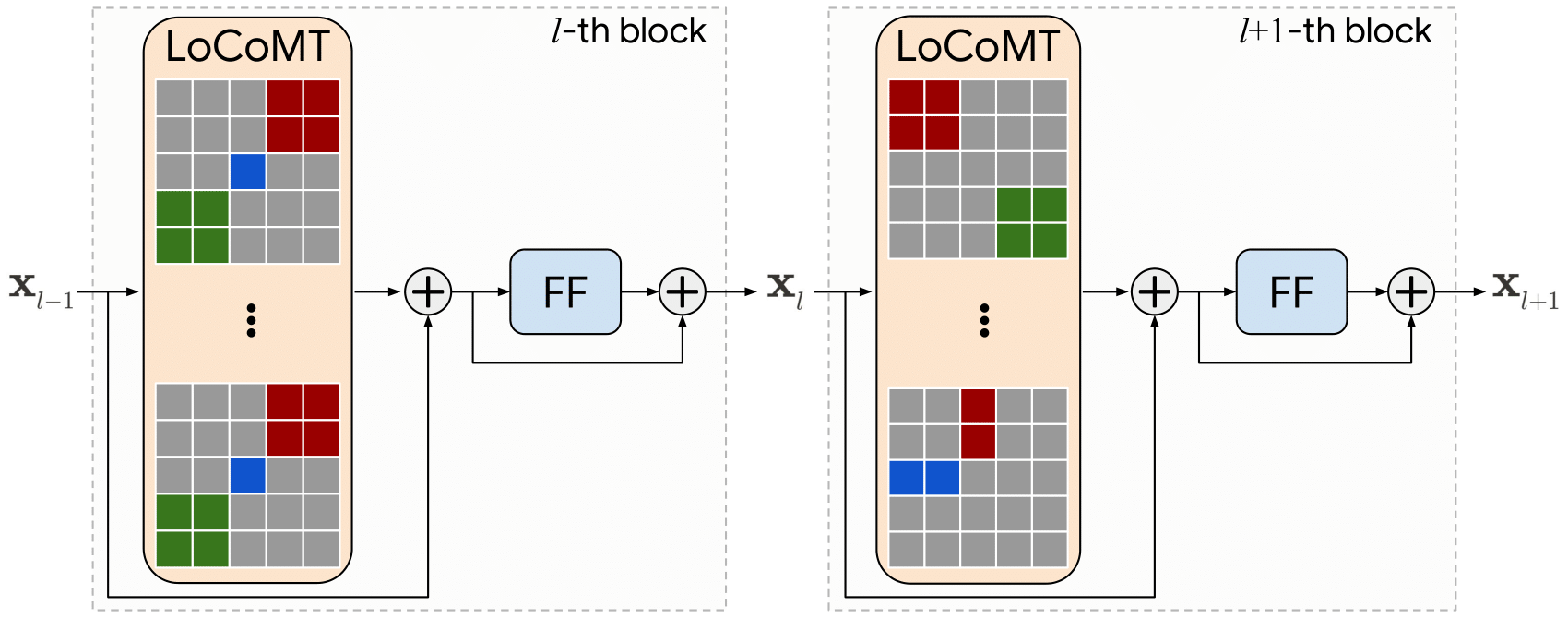}
    \caption{
        Two consecutive Low-Cost Multimodal Transformer (\mname) layers. Our model can assign different attention patterns across the attention heads and layers, allowing for flexible control over multimodal signals.
    }
    \label{fig:eco}
\end{figure}

%% file: contents/03Experiments.tex
\subsection{Experiment Settings}
\par \noindent \textbf{Dataset} 
We train and evaluate our model on two YouTube video classification datasets: Audioset~\cite{audioset} and MedVidCL~\cite{medvidcl}.
We utilized the labeled data available as of January 2023 because some videos became inaccessible after the dataset's release.
This resulted in 18,629 Audioset samples for training and 17,072 for evaluation, and 4,096 MedVidCL samples for training and 1,832 for evaluation.

\par \noindent \textbf{Evaluation Metrics} We report the mean average precision (mAP) for Audioset and both precision and recall for MedVidCL as quantitative performance metrics. Additionally, we report giga floating point operations (GFLOPs) of matrix multiplication to compare efficiency.

\par \noindent \textbf{Models}
In all experiments, multimodal Transformers first encode each modality using $L-L_f$ unimodal layers and then pass the concatenation of them through $L_f$ fusion layers. 
Classification logits are obtained by applying a linear projection to the mean output representation of the [CLS] tokens from each modality.
The fusion layer computes the attention matrix by adopting one of the attention patterns: self-, cross-, multimodal, bottleneck or \mname in Fig.~\ref{fig:pattern}. 
Furthermore, we also tested two sparse attention mechanisms proposed in Longformer~\cite{longformer} and Bigbird~\cite{bigbird} for multimodal fusion.
If the model is made up entirely of fusion layers, we signify the model by ${all}$. 
We refer to the number of attention views assigned to the attention heads as the view frequency $f=(f_1, f_2, \ldots, f_{{}_mC_2+1})$, where $f_i$ is the number of attention heads that $P_i$ is assigned. 
For the experiments, we applied the same view frequency across all fusion layers, except for the experiments in Tab.~\ref{tab:layerwise}. 
Note that we evaluated all possible $L_f$ and $f$ and then reported the best result.

\par \noindent \textbf{Implementation Details}
We used ViT-base/16 trained on ImageNet-21k as a backbone architecture. 
During training, we sampled frames and spectrograms from a randomly selected temporal crop.
For inference, we sampled four temporal segments with a uniform stride, averaging logits calculated from these segments.
From the 8-second temporal crop, we extracted eight $224\times224$ RGB frames and a $128\times800$ log-mel spectrogram. 
Specifically, the spectrogram was computed using a 25 ms Hamming window with a hop length of 10 ms from 16kHz audio.
 We split both RGB frames and the spectrogram into $16\times16$ non-overlapping patches and flattened them into 768-dimensional vectors.
For text, we used the BERT~\cite{bert} tokenizer and embedding.
We applied the data augmentation used in AST~\cite{ast} for audio and ViViT~\cite{vivit} for video.
We trained the model using cross-entropy loss for 20 epochs and with a batch size of 64. 
We used a synchronous SGD optimizer with a momentum of 0.9 following a cosine learning rate schedule with a linear warm-up.

\input{tables/01experiments}

\subsection{Results}
As shown in Tab.~\ref{tab:minias}, \mname demonstrated performance comparable to the best models. 
Furthermore, \mname reduced GFLOPs by 6.2\% on Audioset and by a significant 51.3\% on MedVidCL, compared to the best-performing model of each dataset. 
This suggests that \mname can reduce computational cost with only a minimal decrease in performance. 
In contrast, MBT fell behind Cross in performance and required more computational resources in MedVidCL. 
This may be due to bottleneck fusion that compresses multimodal signals into a small number of bottleneck tokens, potentially resulting in loss of information. 
Both Longformer and Bigbird show a less favorable performance-cost trade-off than \mname in all datasets, underscoring the limitation of traditional sparse attention mechanisms in handling multimodal input.

\input{figures/ablations}

\subsection{Analysis}
\label{sec:analysis}
\par \noindent \textbf{Performance-Cost Trade-off}
We investigate how much computational cost we can save with a negligible reduction in performance.
To this end, we trained \mname on MedVidCL while changing the view frequency and the number of fusion layers that affect both the F1 score and the GFLOPs.
We also examined the F1 score and GFLOPs of Multi and MBT while varying the number of fusion layers.
Fig.~\ref{fig:ablations}a illustrates the trade-off between the F1 score and the GFLOPs of the models.
We observed that \mname reduced the GFLOPs by 81\%, 43.2\%, and 31.5\% compared to Self, Multi, and MBT, respectively, while losing only 1\% of the F1 score.
We note that, despite similar computational costs, the models demonstrate varying performance levels depending on the configuration of the attention view.
It suggests that the proper combination of the number of fusion layers and the view frequency can compensate for the decrease in the capacity of multi-head attention.

\par \noindent \textbf{Effect of Fusion Layers}
As is known from previous studies~\cite{fusionsurvey}, the level of fusion has a great effect on the performance of multimodal models.
Furthermore, the number of fusion layers also affects the computational cost of \mname.
To analyze the effect of the level of fusion, we measured the performance against the number of fusion layers on Audioset as shown in Fig.~\ref{fig:ablations}b.
We also evaluated the impact of the view frequency and illustrated with a different color.
We found that multimodal fusion at lower-level features achieved a worse mAP.
Furthermore, the decision-level fusion (\textit{i.e.,} without fusion layers) performed worse than \mname with a small number of fusion layers.
This implies that constraining cross-modal connections to higher layers in the model enhances the ability of earlier layers to learn unimodal features while continuing to benefit from the multimodal interaction across multiple layers.
We further note that as the number of self-attention heads increases, there is a slower reduction in performance when the number of fusion layers increases.
The larger number of self-attention heads increases the capacity to model the correlation between inputs belonging to the same modality, so it may replace some functionalities of unimodal encoders.

\par \noindent \textbf{Effect of View Frequency}
We fixed the number of fusion layers at 4 and measured mAP and GFLOPs against the view frequency on Audioset.
Here, we refer to the number of self-attention heads as the view frequency because the set of patterns consists of one self- and one cross-attention. 
We illustrate the result in Fig.~\ref{fig:ablations}c.
We observed that performance always improved when mixing heads with different patterns, rather than constructing all heads as a single pattern.
Furthermore, the performance of \mname increased as the number of self-attention heads increased.
However, the GFLOPs increased in proportion to the number of self-attention heads, while the performance did not.
This implies that we can improve the performance-cost trade-off by adjusting the view frequency.

\par \noindent \textbf{Layer-wise Assignment of View Frequency}
Since \mname can flexibly control the multimodal signal by assigning a different view frequency to each layer, it enables the optimization of performance-cost trade-offs. We assessed four allocation strategies—\textit{spread, bottleneck, alternating,} and \textit{random}—for view frequencies. Both \textit{spread} and \textit{bottleneck} strategies progressively decrease the number of self-attention heads by $\lfloor\frac{n_h}{L_f}\rfloor$ in the first $\frac{L_f}{2}$ layers. The \textit{spread} strategy maintains this decrease in subsequent layers, while the \textit{bottleneck} strategy reverses this and increases the number of self-attention heads. Inspired by the cross-modal encoder in LXMERT~\cite{lxmert}, the \textit{alternating} strategy iteratively applies a cross-attention layer following a self-attention layer. To verify the effectiveness of human-crafted attention patterns, we compare them with randomly generated attention patterns. For this, we sample view frequencies with five random seeds and report the mean and standard deviation. We evaluated four strategies on Audioset with $L_f=4, 12$ and report mAP in Tab.~\ref{tab:layerwise}. The results show that the \textit{spread} strategy outperform others at $L_f=12$. On the contrary, when applied to high-level features ($L_f=4$), the \textit{spread} strategy fails to integrate the multimodal signal due to an insufficient number of cross-attention heads. The results suggest that allocating more self-attention heads for low-level features and cross-attention heads for high-level features can enhance multimodal understanding. Interestingly, our findings also indicate that the \textit{random} strategy can match or even exceed the performance of human-crafted attention patterns.
This potentially indicates the existence of optimal view frequencies for different $L_f$, and we leave the search algorithm for optimal view frequencies for future work.

\input{tables/04layerwise}

%% file: tables/01experiments.tex
\begin{table}[t]
    \centering
    \resizebox{1.0\columnwidth}{!}{
    \begin{tabular}{l|cc|ccc}
        \hline 
        \multicolumn{1}{c}{Task} & \multicolumn{2}{|c|}{\textbf{Audioset}} & \multicolumn{3}{c}{\textbf{MedVidCL}}\\
        \hline 
        \multicolumn{1}{c|}{Model} & \textbf{mAP} & \textbf{GFLOPs} & \textbf{Precision} & \textbf{Recall} & \textbf{GFLOPs}\\
        \hline 
        \textit{unimodal} & & & & & \\
        \quad ViT~\cite{vit} & 23.1 & 45.3 & 81.1 & 79.9 & 45.3\\
        \quad AST~\cite{ast} & 28.2 & 2.9 & 52.9 & 50.7 & 2.9\\
        \quad BERT~\cite{bert} & - & - & 80.4 & 79.3 & 0.1\\
        \hline 
        \textit{multimodal} & & & & & \\
        \quad Self$_{all}$ & 37.8 & 48.2 & 77.2 & 66.4 & 48.3 \\
        \quad Multi$_{all}$ & 35.1 & 71.4 & 82.5 & 81.6 & 76.1  \\
        \quad Multi  & 39.0 & 52.1 & 81.8 & 80.9 & 62.2  \\
        \quad Cross$_{all}$ & 25.9 & 23.2 & 71.8 & 69.4 & 27.8\\
        \quad Cross & 38.2 & 39.9 & 81.5 & 80.5 & 44.8 \\
        \quad MBT~\cite{mbt} & 39.6 & 48.4 & 81.1 & 80.3 & 48.4 \\
        \quad Longformer~\cite{longformer} & 33.0 & 41.8 & 80.3 & 79.4 & 41.8\\
        \quad Bigbird~\cite{bigbird} & 38.8 & 45.8 & 81.4 & 80.6 & 46.0\\
        \quad \mname & 39.4 & 45.4 & 82.0 & 81.1 & 36.8  \\
        \hline 
    \end{tabular}
}
    \caption{Test-set results on Audioset and MedVidCL.}
    \label{tab:minias}
\end{table}

%% file: figures/ablations.tex
\begin{figure*}[t]
    \centering
    \includegraphics[width=2.0\columnwidth]{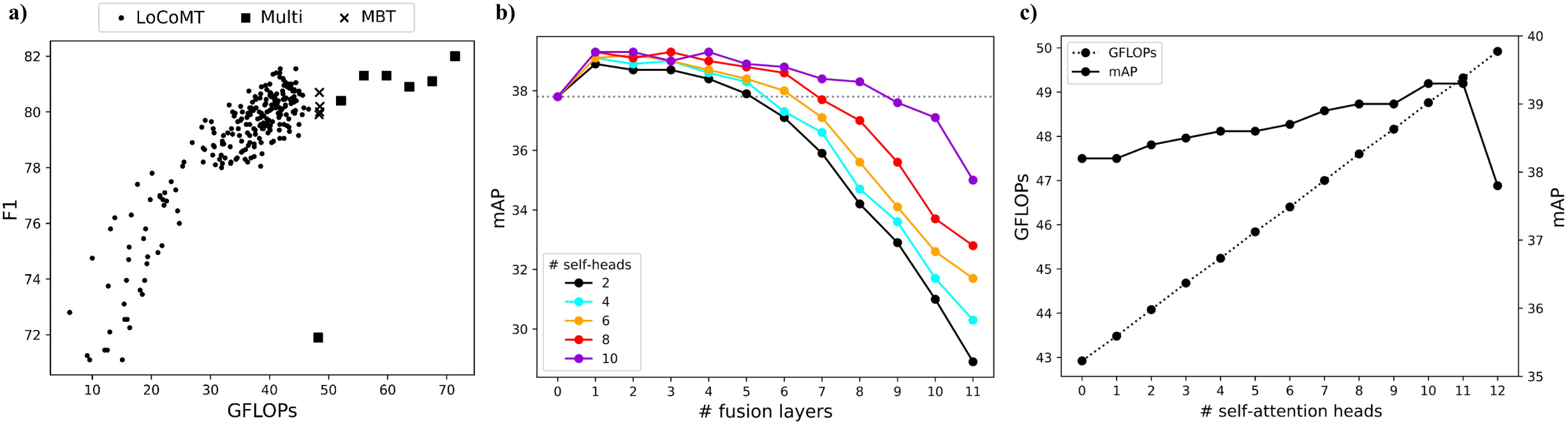}
    \caption{a) The performance-efficiency trade-off. b) Effect of varying the number of fusion layers and the view frequency. c) Effect of varying the view frequency while keeping the number of fusion layers constant.}
    \label{fig:ablations}
\end{figure*}

%% file: tables/04layerwise.tex
\begin{table}[t]
    \centering
    \resizebox{0.5\columnwidth}{!}{%
    \begin{tabular}{l|c|c}
        \hline 
        \textbf{Strategy} & $L_f=4$ & $L_f=12$\\
        \hline
         spread & 38.5 & 37.2\\
         alternating & 38.1 & 31.6\\
         bottleneck & 39.0 & 27.4\\
         random & 39.2 (0.2) & 33.7 (3.5)\\
        \hline 
    \end{tabular}
}
    \caption{Test-set results of four strategies on Audioset.}
    \label{tab:layerwise}
\end{table}

%% file: contents/04Conclusion.tex
We propose Low-Cost Multimodal Transformer (\mname), a novel attention mechanism designed for efficient multimodal fusion. 
Through our experiments, we show that \mname matches or outperforms existing multimodal Transformers in classification tasks, while requiring fewer operations. 
Moreover, we have found that certain combinations of attention heads with different functionalities can significantly reduce the computational cost with only a minimal drop in performance. 
A limitation of \mname is the need to adjust the view frequency to achieve the optimal balance between performance and cost.
However, as shown in Tab.~\ref{tab:layerwise}, even a random strategy outperforms other strategies at $L=4$. 
This result suggests that suboptimal configurations of attention views can achieve satisfactory performance. 
For future work, we plan to develop a search algorithm for the optimal configuration of views.